\documentclass[11pt,a4paper]{article}
\usepackage[hyperref]{acl2017}
\usepackage{times}
\usepackage{latexsym}
\usepackage{tikz}
\usepackage{graphicx}
\usepackage{hyperref}
\usepackage{amsmath}
\usepackage{float}
\usepackage{amsfonts}
\usepackage{comment}
\usepackage{afterpage}
\usepackage{multirow}
\usepackage{array}
\usepackage{tabularx}
\usepackage{url}
\newcolumntype{L}[1]{>{\raggedright\let\newline\\\arraybackslash\hspace{0pt}}m{#1}}

\usepackage{parallel,enumitem}
\usepackage{subcaption}
\usepackage[font=small,labelfont=bf]{caption}

\aclfinalcopy 

\title{Siamese Neural Networks with Random Forest for detecting duplicate question pairs}

\author{Ameya Godbole \textsuperscript{1}, Aman Dalmia \textsuperscript{1}, Sunil Kumar Sahu \textsuperscript{2},
\\  \textsuperscript{1} Department of Electronics and Communication Engineering
\\  \textsuperscript{2} Department of Computer Science and Engineering
\\ Indian Institute of Technology Guwahati, Assam, India 
\mbox{}\\
{\tt \{godbole, a.dalmia, sunil.sahu\}@iitg.ernet.in}\\
}

\date{}

\begin{document}
\tikzstyle{rect} = [draw, rectangle, fill=white!20,minimum width = 1cm, text centered, minimum height=1cm]
\tikzstyle{circ} = [draw, circle, fill=white!20, minimum width=8pt, inner sep=3pt]
\tikzstyle{arrow} = [thick,->,>=stealth]
\maketitle
\begin{abstract}
Determining whether two given questions are semantically similar is a fairly challenging task given the different structures and forms that the questions can take. In this paper, we use Gated Recurrent Units(GRU) in combination with other highly used machine learning algorithms like Random Forest, Adaboost and SVM for the similarity prediction task on a dataset released by Quora, consisting of about 400k labeled question pairs. We got the best result by using the Siamese adaptation of a Bidirectional GRU with a Random Forest classifier, which landed us among the top 24\% in the competition Quora Question Pairs hosted on Kaggle.
\end{abstract}
\section{Introduction}
Understanding text is an important task and identifying semantic similarity between sentences finds various applications in machine translation, sentiment analysis, natural language understanding, etc. Sentences tend to have complex structure, may include unforeseen modifications like synonyms, acronyms and spelling mistakes and mostly have varying lengths. This makes the task of detecting duplicate questions significantly hard. We use the definition of duplicate questions \cite{hommadetecting} as questions which can be answered by the exact same answers. Q\&A forums like Quora encounter many situations where different people have asked semantically similar questions. To correctly identify such questions and merge them into a single canonical question would greatly improve their efficiency as well as user experience as readers won't have to look at multiple places for answers, thus getting answers faster and writers won't have to write the same answer to multiple versions of the same question.

A simple duplicate detection approach is a word-based comparison where we can find the word-based similarity between the two questions using standard information retrieval measures like tfidf and then classify question pairs with similarity scores above a certain threshold as duplicates. But this approach ignores the semantic meaning in the questions and hence, we need more intelligent approaches than these simple measures. Recent advances in deep learning specifically GRUs, which are a modification to the traditional Recurrent Neural Networks (RNN), have made significant gains in tasks like semantic equivalence detection, surpassing traditional machine learning techniques that use hand-picked features.

In this paper, we demonstrate one such approach using GRUs in combination with other machine learning algorithms that are widely used for the similarity detection task like Random Forests and SVMs.

\section{Related Work}
Because of its wide variety of applications, the problem of detecting semantically similar sentences is being studied for a long time. Early work to quantify the similarity between sentences used manually engineered features like word overlap \cite{broder1997resemblance} along with traditional machine learning algorithms like Support Vector Machines (SVMs) \cite{dey2016paraphrase}.

Since the resurgence in Deep learning, Neural Network approaches have been the state-of-the-art in a wide range of NLP tasks.CNNs have shown good results in sentiment analysis tasks whereas using pre-trained word vectors derived using \textit{word2vec} have been performing well over sentence classification tasks. However, most of the deep learning methods proposed for detecting semantic similarity have used a Siamese Neural Network architecture, which uses the same neural network as a feature extractor for both the sentences and then compares them using a distance metric. 

Our model is based on the ensemble approach to take the best of both worlds by combining the similarity score from the GRU with a few hand-engineered features and using it along with various successful machine learning algorithms.
\begin{center}
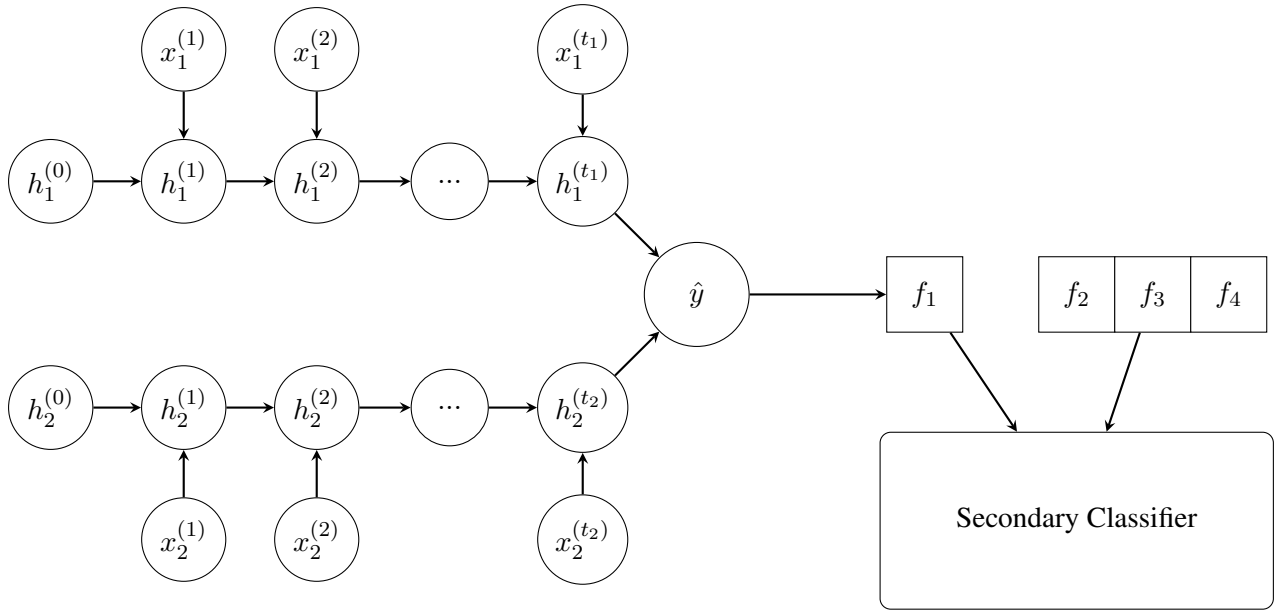
\begin{figure*}[!htb]
\centering
\begin{tikzpicture}{node distance = 1cm, auto}
\node [circ] (h0) {$h_1^{(0)}$};
\node [circ, right of=h0, node distance = 1.75cm] (h1) {$h_1^{(1)}$};
\node [circ, right of=h1, node distance = 1.75cm] (h2) {$h_1^{(2)}$};
\node [circ, right of=h2, inner sep=7.5pt, node distance = 1.75cm] (hdot) {$...$};
\node [circ, right of=hdot, node distance = 1.75cm] (ht1) {$h_1^{(t_1)}$};
\node [circ, above of=h1, node distance = 1.75cm] (x1) {$x_1^{(1)}$};
\node [circ, above of=h2, node distance = 1.75cm] (x2) {$x_1^{(2)}$};
\node [circ, above of=ht1, node distance = 1.75cm] (xt1) 
{$x_1^{(t_1)}$};

\node [circ, inner sep=10pt, below of=ht1, right of=ht1, node distance = 1.5cm] (y) 
{$\hat{y}$};

\node [rect, right of=y, node distance=3cm] (f1) {$f_1$};

\node [rect, right of=f1, node distance=2cm] (f2) {$f_2$};
\node [rect, right of=f2, node distance=1cm] (f3) {$f_3$};
\node [rect, right of=f3, node distance=1cm] (f4) {$f_4$};

\node [circ, below of=h0, node distance = 3cm] (h20) {$h_2^{(0)}$};
\node [circ, right of=h20, node distance = 1.75cm] (h21) {$h_2^{(1)}$};
\node [circ, right of=h21, node distance = 1.75cm] (h22) {$h_2^{(2)}$};
\node [circ, right of=h22, inner sep=7.5pt, node distance = 1.75cm] (h2dot) {$...$};
\node [circ, right of=h2dot, node distance = 1.75cm] (h2t2) {$h_2^{(t_2)}$};
\node [circ, below of=h21, node distance = 1.75cm] (x21) {$x_2^{(1)}$};
\node [circ, below of=h22, node distance = 1.75cm] (x22) {$x_2^{(2)}$};
\node [circ, below of=h2t2, node distance = 1.75cm] (x2t2) 
{$x_2^{(t_2)}$};

\node [rect, inner sep=1cm, rounded corners, below of=f2, node distance=3cm] (cl) {Secondary Classifier};

\draw [arrow] (h0) -- (h1);
\draw [arrow] (h1) -- (h2);
\draw [arrow] (h2) -- (hdot);
\draw [arrow] (hdot) -- (ht1);
\draw [arrow] (ht1) -- (y);
\draw [arrow] (x1) -- (h1);
\draw [arrow] (x2) -- (h2);
\draw [arrow] (xt1) -- (ht1);
\draw [arrow] (h20) -- (h21);
\draw [arrow] (h21) -- (h22);
\draw [arrow] (h22) -- (h2dot);
\draw [arrow] (h2dot) -- (h2t2);
\draw [arrow] (h2t2) -- (y);
\draw [arrow] (x21) -- (h21);
\draw [arrow] (x22) -- (h22);
\draw [arrow] (x2t2) -- (h2t2);
\draw [arrow] (y) -- (f1);
\draw [arrow] (f3) -- (cl);
\draw [arrow] (f1) -- (cl);
\end{tikzpicture}
\caption{Siamese Neural Network Architecture: $x_1$ and $x_2$ represent the inputs, $h_1$ and $h_2$ represent the hidden states and the predicted score $\hat{y}$ is used as one of the features (the other three are represented by $f_1$, $f_2$ and $f_3$) for the Random Forest classifier that gave the best result.} \label{fig:M}
\end{figure*}
\end{center}

\section{Model Architecture} 

We present our model based on bidirectional RNN as shown in Figure \ref{fig:M} for the semantic similarity detection task. Our model uses embedding features of words in the input layer and learns higher level representations in the subsequent layers and makes use of those higher level features to perform the final classification (duplicate or not). We now briefly explain about each component of our model.

\subsection{Embedding or Lookup Layer}
In this layer every input feature is mapped to a dense feature vector. Let us say that $E_w$ be the embedding matrix of W where W represents the words in the vocabulary.
The $E_w$ $\in \mathbb{R}^{n_w \times d_w}$ embedding matrix is initialized to the 50-dimensional GloVe vectors pre-trained by \cite{pennington2014glove} on the Wikipedia 2014 and Gigaword 5 datasets. Here $n_w$ refers to length of the word dictionary and $d_w$ refers to dimension of word embedding.

\subsection{Bidirectional RNN Layer}
RNN is a powerful model for learning features from sequential data. We use both the GRU \citep{chung2014empirical} variant of RNN in our experiments as they handle the vanishing and exploding gradient problem \citep{pascanu2012understanding} in a better way. We use bidirectional version of RNN \citep{Graves13} where for every word forward RNN captures features from the past and the backward RNN captures features from future, inherently each word has information about whole sentence.

\subsection{Feed Forward Neural Network}
The hidden state of the bidirectional RNN layer acts as sentence-level feature ($g$), the word and entity type embeddings ($l$) act as a word-level features, are both concatenated and passed through a series of hidden layers,  with dropout \citep{srivastava2014dropout} and an output layer. In the output layer, we use $Sigmoid$ activation function to obtain probability score for the question pair being duplicate.

\subsection{Training and Hyperparameters}
We use cross entropy loss function and the model is trained using the Adam optimizer. The implementation\footnote{Implementation is available at \url{https://github.com/dalmia/Quora-Question-Pairs}} of the model is done in python language using $TensorFlow$ \citep{abadi2016tensorflow} library.

We use training and development set for hyperparameter selection. We use word embeddings of $50$ dimension, entity, $3$-layer stacked RNN with hidden state dimensions of $250$, $500$, $250$ respectively and $2$ hidden layers with dimension $1000$ and $1024$. We use dropout in the first hidden layer with a \textit{keep\_prob} of $0.8$.

\section{Experiments and discussion}
\label{sec:exp}
\subsection{Dataset Preprocessing}
\label{sec:dataset}
We use the preprocessing procedure of \cite{hommadetecting} with some changes to reflect the dataset for the competition. We performed sentence tokenization using the Stanford Tokenizer \footnote{https://nlp.stanford.edu/software/tokenizer.html} with all characters in lowercase and \textit{ptb3Escaping} disabled. Multiprocessing was used to speed up the procedure. Special token 'UNK' was used for words not found in the GloVe embeddings vocabulary. \cite{hommadetecting} suggest using 30 as the standardized sentence length to allow batching of data during training. They suggested this after hyperparameter search. Instead, we decided to set sentence length to 40 words by observing the distribution of the length of tokenized sentences. Shorter sentences are padded at the beginning by special zero-padding token and longer sentences are truncated. We initialize the word embeddings to the 50-dimensional GloVe vectors.

\begin{figure*}[!h]
\begin{center}
\includegraphics[width=0.99\textwidth]{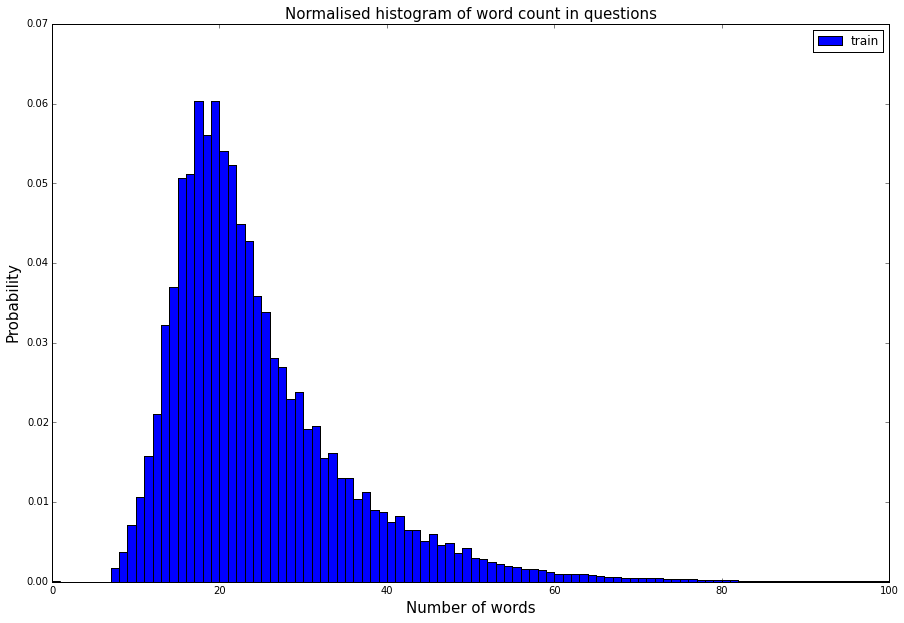}
\caption{Histogram of sentence lengths after tokenization in the training set}
\label{fig:word_dist}
\end{center}
\end{figure*}

\subsection{Dataset Augmentation}
\label{sec:dataset}
We use data augmentation as suggested by \cite{hommadetecting} for training the Siamese-GRU model. The first type of augmentation is necessary for improving the response of the Siamese arms to flipped input. Accordingly, the dataset is augmented by flipping the question pairs. Additional positive samples are generated by pairing each unique question with itself. Finally, negative samples are generated to balance the augmented dataset by randomly selecting 2 questions that do not have common known duplicates. The procedure resulted in 3x more training samples (see Table \ref{table:augment}). This augmented data was not used for training the secondary classifier.

\begin{table}[h]
\resizebox{\columnwidth}{!}{%
\begin{tabular}{|c|c|c|}
\hline
\textbf{Data} & \textbf{Before} & \textbf{After}\\
\hline
Positive & 149263 & 836459\\
\hline
Negative & 255027 & 836454\\
\hline
Total samples & 404290 & 1672913\\
\hline
Ratio (pos:neg) & 0.58 & 1.0\\
\hline
\end{tabular}
}
\caption{Dataset augmentation details}
\label{table:augment}
\end{table}

\subsection{Additional Features}
\label{feat}
Rather than using the base Siamese-GRU model as the final classifier, we combined the model output with 3 additional features which was used as the input to a secondary classifier. These features were suggested on Kaggle's discussion forum. Word match share expresses the overlap between the words of the 2 sentences under consideration. Another feature is obtained by weighing the common words by their TF-IDF (term-frequency-inverse-document-frequency) weight while measuring overlap \footnote{https://www.kaggle.com/anokas/data-analysis-xgboost-starter-0-35460-lb}. The third feature for every pair is the number of questions that are known to be duplicates \footnote{https://www.kaggle.com/tour1st/magic-feature-v2-0-045-gain}. A non-zero value for this feature is a definite indicator of duplicates because if two questions have a common question marked duplicate of both, by transitivity the questions under consideration are duplicates.

\subsection{Secondary Classifiers}
The base Siamese-GRU model is used in conjunction with the 3 features described above as input for a secondary classifier. We experimented with AdaBoost, SVM and Random Forest classifiers. The best results were obtained by the Random Forest with suitable regularization in the form of restriction of maximum tree depth. The model performance is outlined in Table \ref{table:performance}. The score is calculated by Kaggle's engine using cross-entropy loss. The model nomenclature states the number of stacked GRU layers followed by the number of fully-connected layers leading to the similarity node. Dropout has been used between fully-connected layers where mentioned.

\begin{table}[h]
\resizebox{\columnwidth}{!}{%
\begin{tabular}{|c|c|}
\hline
\textbf{Method} & \textbf{Cross-Entropy loss}\\
\hline
GRU\_1\_1 + SVM & 0.27761\\
\hline
GRU\_1\_1 + AdaBoost & 0.64782\\
\hline
\textbf{GRU\_1\_1 + RF} & \textbf{0.24365}\\
\hline
GRU\_3\_1 + AdaBoost & 0.65615\\
\hline
GRU\_3\_1 + RF & 0.24519\\
\hline
GRU\_3\_2 + AdaBoost & 0.63288\\
\hline
GRU\_3\_2 + RF & 0.24544\\
\hline
GRU\_3\_2 + Dropout + AdaBoost & 0.63305\\
\hline
GRU\_3\_2 + Dropout + RF & 0.29568\\
\hline
\end{tabular}
}
\caption{Comparison of performance of our models}
\label{table:performance}
\end{table}

\section{Conclusion and Future Work}
In this paper we have demonstrated an approach for detecting semantic similarity by learning higher level features using RNN. Our experiments have re-affirmed the usefulness of ensembling and landed us in the top 25 percentile band. In future we would like to extend our experiments to other models such as ABCNN (\citet{DBLP:journals/corr/YinSXZ15}) and ESIM (\citet{DBLP:journals/corr/ChenZLWJ16}) which have shown better performance than traditional NLP techniques.

\bibliography{acl2017}

\begin{thebibliography}{}
\expandafter\ifx\csname natexlab\endcsname\relax\def\natexlab#1{#1}\fi

\bibitem[{Abadi et~al.(2016)Abadi, Agarwal, Barham, Brevdo, Chen, Citro,
  Corrado, Davis, Dean, Devin et~al.}]{abadi2016tensorflow}
Mart{\'\i}n Abadi, Ashish Agarwal, Paul Barham, Eugene Brevdo, Zhifeng Chen,
  Craig Citro, Greg~S Corrado, Andy Davis, Jeffrey Dean, Matthieu Devin, et~al.
  2016.
\newblock Tensorflow: Large-scale machine learning on heterogeneous distributed
  systems.
\newblock {\em arXiv preprint arXiv:1603.04467\/} .

\bibitem[{Broder(1997)}]{broder1997resemblance}
Andrei~Z Broder. 1997.
\newblock On the resemblance and containment of documents.
\newblock In {\em Compression and Complexity of Sequences 1997. Proceedings\/}.
  IEEE, pages 21--29.

\bibitem[{Chen et~al.(2016)Chen, Zhu, Ling, Wei, and
  Jiang}]{DBLP:journals/corr/ChenZLWJ16}
Qian Chen, Xiaodan Zhu, Zhen{-}Hua Ling, Si~Wei, and Hui Jiang. 2016.
\newblock \href{http://arxiv.org/abs/1609.06038}{Enhancing and combining
  sequential and tree {LSTM} for natural language inference}.
\newblock {\em CoRR\/} abs/1609.06038.
\newblock
  \href{http://arxiv.org/abs/1609.06038}{http://arxiv.org/abs/1609.06038}.

\bibitem[{Chung et~al.(2014)Chung, G{\"{u}}l{\c{c}}ehre, Cho, and
  Bengio}]{chung2014empirical}
Junyoung Chung, {\c{C}}aglar G{\"{u}}l{\c{c}}ehre, KyungHyun Cho, and Yoshua
  Bengio. 2014.
\newblock Empirical evaluation of gated recurrent neural networks on sequence
  modeling.
\newblock {\em CoRR\/} abs/1412.3555.

\bibitem[{Dey et~al.(2016)Dey, Shrivastava, and Kaushik}]{dey2016paraphrase}
Kuntal Dey, Ritvik Shrivastava, and Saroj Kaushik. 2016.
\newblock A paraphrase and semantic similarity detection system for user
  generated short-text content on microblogs.
\newblock In {\em COLING\/}. pages 2880--2890.

\bibitem[{Graves(2013)}]{Graves13}
Alex Graves. 2013.
\newblock Generating sequences with recurrent neural networks.
\newblock {\em CoRR\/} abs/1308.0850.

\bibitem[{Homma et~al.()Homma, Sy, and Yeh}]{hommadetecting}
Yushi Homma, Stuart Sy, and Christopher Yeh. ????
\newblock Detecting duplicate questions with deep learning .

\bibitem[{Pascanu et~al.(2012)Pascanu, Mikolov, and
  Bengio}]{pascanu2012understanding}
Razvan Pascanu, Tomas Mikolov, and Yoshua Bengio. 2012.
\newblock Understanding the exploding gradient problem.
\newblock {\em CoRR, abs/1211.5063\/} .

\bibitem[{Pennington et~al.(2014)Pennington, Socher, and
  Manning}]{pennington2014glove}
Jeffrey Pennington, Richard Socher, and Christopher~D Manning. 2014.
\newblock Glove: Global vectors for word representation.
\newblock In {\em EMNLP\/}. volume~14, pages 1532--1543.

\bibitem[{Srivastava et~al.(2014)Srivastava, Hinton, Krizhevsky, Sutskever, and
  Salakhutdinov}]{srivastava2014dropout}
Nitish Srivastava, Geoffrey~E Hinton, Alex Krizhevsky, Ilya Sutskever, and
  Ruslan Salakhutdinov. 2014.
\newblock Dropout: a simple way to prevent neural networks from overfitting.
\newblock {\em Journal of Machine Learning Research\/} 15(1):1929--1958.

\bibitem[{Yin et~al.(2015)Yin, Sch{\"{u}}tze, Xiang, and
  Zhou}]{DBLP:journals/corr/YinSXZ15}
Wenpeng Yin, Hinrich Sch{\"{u}}tze, Bing Xiang, and Bowen Zhou. 2015.
\newblock \href{http://arxiv.org/abs/1512.05193}{{ABCNN:} attention-based
  convolutional neural network for modeling sentence pairs}.
\newblock {\em CoRR\/} abs/1512.05193.
\newblock
  \href{http://arxiv.org/abs/1512.05193}{http://arxiv.org/abs/1512.05193}.

\end{thebibliography}
\bibliographystyle{acl_natbib}

\appendix

\end{document}